\documentclass[letterpaper, 10 pt, conference]{ieeeconf}
\IEEEoverridecommandlockouts
\usepackage{graphicx}
\usepackage{amsfonts}
\usepackage{cuted}
\usepackage{tabularray}
\usepackage{multirow}
\usepackage{graphicx}
\usepackage{cite} 
\usepackage{setspace} 
\usepackage[utf8]{inputenc}
\usepackage{textcomp}
\usepackage{graphicx} 
\usepackage{amsmath,amssymb}  
\usepackage{bm}  
\usepackage{mathtools}
\usepackage{physics}
\usepackage{memhfixc} 
\usepackage{pdfsync}  
\usepackage{fancyhdr}
\usepackage{xcolor} 
\usepackage{subfigure}
\usepackage{algorithm}
\usepackage{algorithmic}
\usepackage{amsfonts}
\usepackage{times}
\usepackage{epstopdf}
\usepackage{multirow}
\usepackage{booktabs}
\usepackage{caption}
\usepackage{lipsum}
\usepackage{hyperref} 
\usepackage{gensymb}
\usepackage{arydshln}
\usepackage{cleveref}
\usepackage{adjustbox}
\usepackage{afterpage}

\title{\LARGE \bf
SPOTS: Stable Placement of Objects with Reasoning in Semi-Autonomous Teleoperation Systems
}

\author{{Joonhyung~Lee$^{1}$,
        Sangbeom~Park$^{1}$,
        Jeongeun~Park$^{1}$,
        Kyungjae~Lee$^{2}$,
        and~Sungjoon~Choi$^{1}$,}% <-this % stops a space
\thanks{$^{1}$Joonhyung Lee, Sangbeom~Park, Jeongeun~Park and Sungjoon~Choi are with the Department of Artificial Intelligence, Korea University, Seoul, Korea
(email: {\tt\small \{dlwnsgud8823, sangbeom-park, baro0906, sungjoon-choi\}@korea.ac.kr})}%
\thanks{$^{2}$Kyungjae Lee is with the Department of Artificial Intelligence, Chungang University, Seoul, Korea 
        (email: {\tt\small kyungjae.lee@ai.cau.ac.kr})} %
}

\begin{document}
\maketitle
 \thispagestyle{empty}
\pagestyle{empty}

%
% Abstract
%
\begin{abstract}
Pick-and-place is one of the fundamental tasks in robotics research. 
However, the attention has been mostly focused on the ``pick'' task, leaving the ``place'' task relatively unexplored. 
In this paper, we address the problem of placing objects in the context of a teleoperation framework. 
Particularly, we focus on two aspects of the place task: stability robustness and contextual reasonableness of object placements. 
Our proposed method combines simulation-driven physical stability verification via real-to-sim
and the semantic reasoning capability of large language models. 
In other words, given place context information (e.g., user preferences, object to place, and current scene information), our proposed method outputs a probability distribution over the possible placement candidates, considering the robustness and reasonableness of the place task. 
Our proposed method is extensively evaluated in two simulation and one real world environments and we show that our method can greatly increase the physical plausibility of the placement as well as contextual soundness while considering user preferences. 
\end{abstract}

%
% Introduction
%
\section{Introduction}

% Framework
\begin{figure*}[h!]
 \centering
 \includegraphics[width=0.95\textwidth]{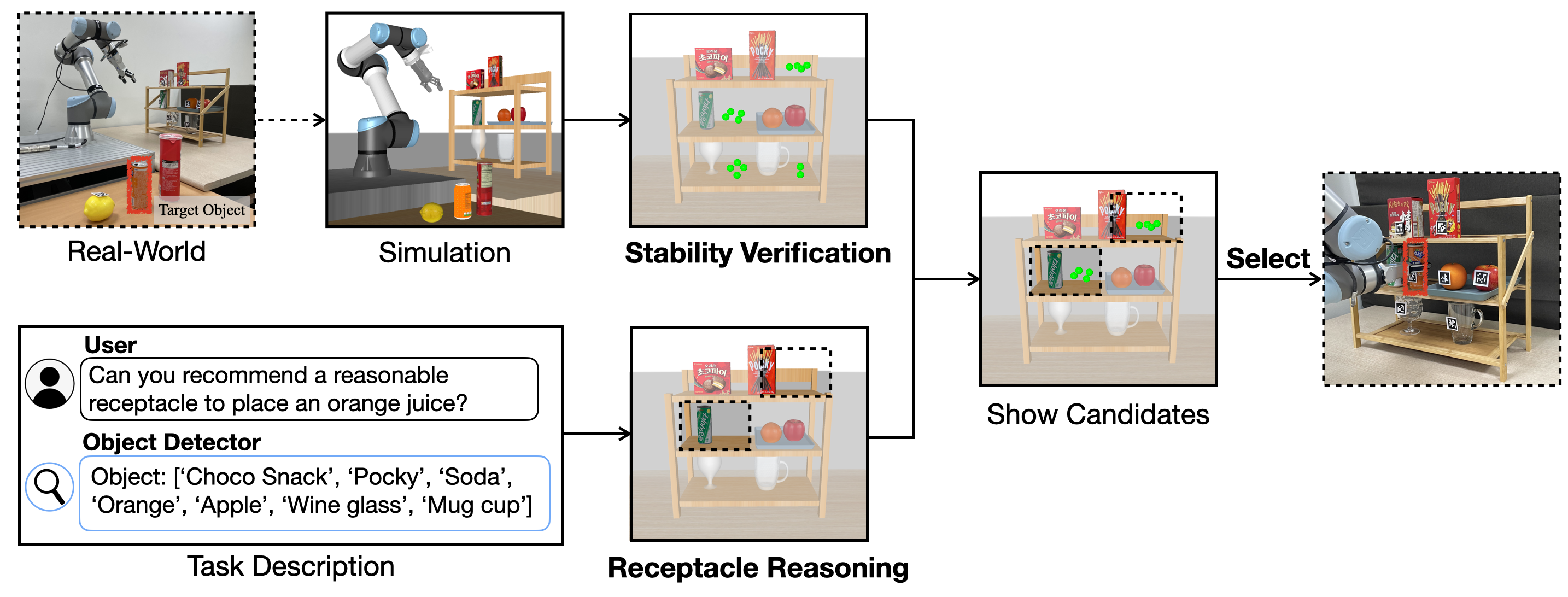}
    \caption{Overall pipeline of the proposed teleoperation framework. With the scene input, the system checks the physical stability of the correct placement. Then, the system verifies the contextually reasonable positions based on the receptacle reasoning step, considering the scene's context, and recommends the coordinates obtained from both processes to the user.}
 \label{fig:overview}
\end{figure*}

Providing a small number of effective options to users in an autonomous system (i.e., a semi-autonomous teleoperation system) has been actively studied~\cite{losey_20_control-latent-space,losey_22_learning-latent-actions,park_22_semi-teleop} to mitigate possible malfunctioning of a fully autonomous system. However, these options often fall short of considering the situational context.
In this paper, we focus on the task of placing objects considering two different aspects, physical robustness and reasonableness considering contextual information, within a semi-autonomous teleoperation framework to provide effective place candidates to a user. 

To achieve robust placement, we utilize simulators to forward simulate multiple place candidates by assuming that the simulation environment is reconstructed from the real world observations (i.e., real-to-sim). 
Contextual soundness is mostly achieved via the reasoning capabilities of large language models (LLMs).
To be specific, scene information (e.g., object to place and other objects in the scene) and optional human preference (e.g., ``I would like to sort objects based on colors.'') are fed into LLMs via vision-language models to restrict the possible place locations. 

While the task of picking and placing objects is a fundamental task, more research focus has been concentrated on picking objects \cite{wang22_efficient, he23_pick2place}. 
However, we argue that the place task is also an essential problem to consider, especially when it comes to semi-autonomous teleoperation (e.g., recommending users a set of appropriate place locations). 
Unlike the picking task, where the target object to pick is usually predetermined, there may exist multiple possible placeable locations, and this makes it more attractive for the semi-autonomous teleoperation framework to be utilized. 

We mainly focus on two aspects of the place task: physical stability and contextual soundness of the placement. 
The former is achieved via simulation-driven verification combined with real-to-sim. 
In other words, given the current scene, we first reconstruct the scene using a physics-based simulator (e.g., MuJoCo~\cite{mujoco}) and check the stability by solving forward dynamics with additional perturbation of the object poses. 
The latter uses the reasoning performance of large language models (LLMs). In particular, the current scene and the object to be placed are described with languages utilizing vision-language models (e.g., OWL-ViT~\cite{owl_vit}). 
Then, the place candidates that passed the physical stability check are examined using the LLM by simply adding the prompt describing the current place candidate to the scene-describing prompt. 

In summary, we made two key contributions to this paper. 
We present \textbf{S}table \textbf{P}lacement of \textbf{O}bjects with reasoning in semi-autonomous \textbf{T}eleoperation \textbf{S}ystems (\textbf{SPOTS}), an approach to a semi-autonomous teleoperation framework that focuses on verifying placement positions with 1) a \textbf{Stability Verification} (i.e., physics-based simulation) step and 2) \textbf{Receptacle Reasoning} (i.e., common knowledge) step by utilizing LLMs that understand scene contexts and reason about the corresponding task without learning.

%
% Related Work
%
\section{Related Work}

\subsection{Semi-Autonomous Teleoperation}
Recently, research on semi-autonomous teleoperation has been actively conducted to minimize human burden while effectively teleoperating a robot. 
In \cite{losey_20_control-latent-space, losey_22_learning-latent-actions}, Losey et al. have addressed a framework to reduce the dimensionality of control inputs for efficient control of high-dimensional robots by embedding high-dimensional trajectories to a low DoF latent actions. However, it uses kinesthetic demonstrations to show the robot arm how to perform a variety of tasks. 
Park et al.~\cite{park_22_semi-teleop} have performed a non-prehensile manipulation task using a high-level option via reinforcement learning (RL) (e.g., pushing away obstacles in a cluttered environment). However, training a policy to handle a complex task, such as placing a book on a bookshelf, is exceedingly challenging. 
Acknowledging these limitations, our method provides various candidates for where to place in a given scene, which is validated in the physics-based simulation environment and fits with common sense using LLMs \cite{ouyang2022training_instruct_gpt, touvron2023llama, chowdhery2022palm}. 

\subsection{Leveraging Simulation for Interactions}
Physics-based simulators have made significant progress in the robotics field \cite{gazebo, pybullet,mujoco,makoviychuk21_isaac}. 
In addition, there have been only a few studies that have explored the translation from real world to simulation environments, in contrast to Sim2Real, with a focus on physical interactions using a physics-based simulator \cite{lim22_real2sim2real, byravan2022nerf2real_mujoco, lv23_samrl, mo2021o2oafford_o2o_afford}.
Lv et al. \cite{lv23_samrl} have conducted a sensing-aware approach to find the robot's next best viewpoint posture in simulation based on the image taken from a specific viewpoint in the real world (i.e., transferring the real world to simulation and updating the posture based on the value function).
Mo et al. \cite{mo2021o2oafford_o2o_afford} proposed a pipeline to leverage differentiable simulation that creates an affordance map by checking the behavior of objects before and after interaction through a simulator when performing a robotic task. 
Similarly, we use the physics-based simulator to validate the interaction of placing objects in the simulator.
This allows us to ensure physical plausibility and verify stability where the target object should be placed.

\subsection{Reasoning capabilities of LLMs for Robotics} 
Alongside the recent success of LLMs~\cite{22_oh,23_openai,23_touvron}, LLMs have been actively studied in robotics. 
Some methods~\cite{ahn_22_saycan, huang_22_inner-monologue} utilize LLMs for task planning a robot, where LLM predicts a sequence of low-level subgoals that are driven by prompt structures. In \cite{CaP, vemprala_23_chatgpt-for-robotics}, they use LLMs to write robot policy codes given language commands by prompting the model with a few demonstrations. LLMs have also been utilized as a reward function for inferring user intentions in negotiation games \cite{kwon_23_reward}, collaborative human-AI interaction games \cite{hu_23_language_human_ai_coordination}, and automating parameterization of reward functions\cite{Language-to-Reward}. 
These methods leverage the semantic priors stored in LLMs to compose new plans or parameterize primitive APIs. 
Mirchandani et al.~\cite{mirchandani_23_general-pattern-machine} have utilized the LLMs as a pattern machine for spatial rearrangement via predicting simple forward dynamics (e.g., moving a red bowl to a green plate).
Inspired by this, we validate the physical plausibility by placing a target object directly in a simulation environment, and we leverage the reasoning capabilities of LLMs to screen out unreliable receptacles considering the spatial relationships (e.g., 6D pose and size of objects) and semantic representations (e.g., color, and flavor).

%
% Problem Formulation
%
\section{Problem Formulation}
\label{Problem Formulation}
We focus on the problem of recommending object place locations to users in the context of semi-autonomous teleoperation. Specifically, we target this issue in complicated and restricted environments with limited free space for placement, such as placing a plate in a dish rack \cite{lee23_opt_dish} or a book on a shelf~\cite{guided-cost-learning, haustein_19_opt_place}.
Given an RGBD image as an input, we aim to predict a set of candidate placement locations that balance the physical stability and reasoning suitability.
Here, physical stability (i.e., robustness) refers to whether an object will stand up as intended without falling over when placed in its space. 
Reasoning suitability (i.e., contextual reasonableness) involves whether the object is placed in a reasonable space.
To accomplish this task, our proposed framework takes as inputs a set of three-dimensional points \( \mathcal{P} = \{ \mathbf{p}_1, \mathbf{p}_2, \ldots, \mathbf{p}_n \} \) in \( \mathbb{R}^3 \) that exist within the robot's accessible workspace, where \( \mathbf{p}_i = (x_i, y_i, z_i) \), as well as a task description that specifies the user's requirements. The framework outputs a density distribution \( \mathcal{D} \) to identify regions where objects can be stably placed.

\begin{figure}[t]
    \centering
    \includegraphics[width=0.45\textwidth]{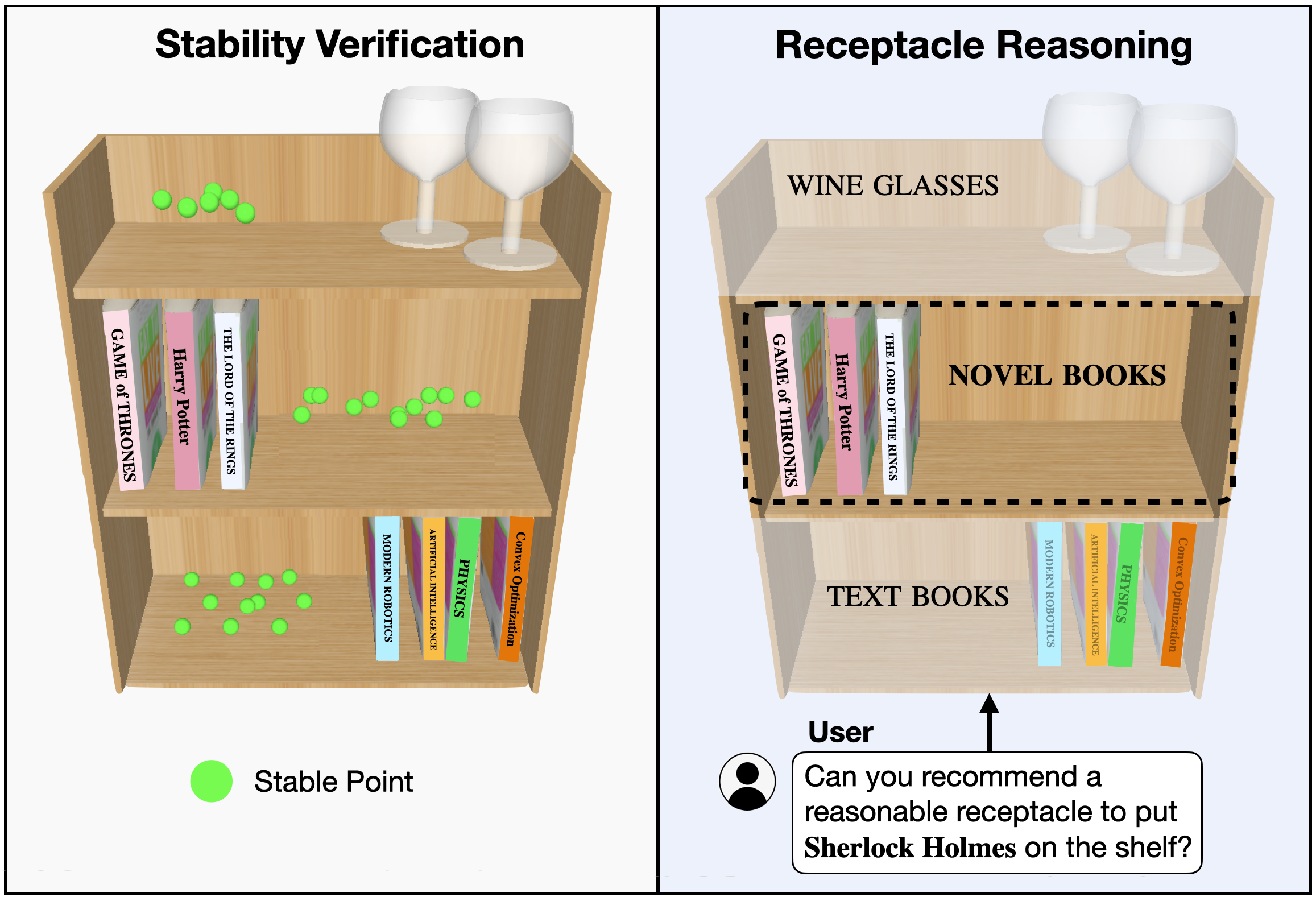}
    \caption{Description of the modules for stability verification and receptacle reasoning. The stability verification module determines if a location is physically stable, and the receptacle reasoning determines the appropriate receptacle.}
\label{fig:overall}
\end{figure}

\section{Proposed Method}
%
% Proposed method
%
In a complex environment, there are a limited number of options to place objects (e.g., placing a plate on a plate rack, placing a book on a bookshelf, or placing an item on a shelf according to its categorization). 
The goal is to find a probability distribution over those coordinates, not just a finite, limited, fixed set of coordinates that can be physically placed.
Our framework has two modules: a) the physics-based screening module (i.e., real-to-sim), which will be introduced in Section \ref{physics-stability}, and b) the reasoning-based suitability check module, which will be illustrated in Section \ref{common-sense-suitability}. 
We would like to emphasize that the robot agent only provides the user with coordinates for the placement of the object, and the user chooses the most appropriate coordinate based on their own needs or preferences.
The overall pipeline is illustrated in Figure \ref{fig:overview}. 

\subsection{Physics-Based Stability Verification}
\label{physics-stability}
The first step of the proposed stability verification is the 3D reconstruction of the real world environment to a physics-based simulator. To accomplish this using an egocentric RGBD image, two key factors must be considered: the objects' spatial and semantic configuration. The spatial arrangements can be measured from depth information based on the camera parameters (i.e., intrinsic matrix and extrinsic matrix). In addition, the semantic attribute of the objects in the scene is captured from object detection models (e.g., Vision Language Models~\cite{owl_vit, kuo2022f_fvlm, li2022lavis_lavis, bravo2023openvocabulary}), which identify and cluster objects in the scene with zero-shot manners. Based on spatial and semantic configuration, we match each object with its corresponding mesh model, which we assume is available. These mesh models are then parsed into the simulator to closely mimic the real world environment. 

We aim to identify regions where objects can be stably placed over a given interaction time \( T \) in simulation.
For each point \( \mathbf{p}_i \) and at each time \( t = 1, 2, \ldots, T \), we have a sequence of scalar components of quaternions belonging to object orientations, denoted as \( q(t, \mathbf{p}_i) \).
We define the set of points \( \mathcal{P}_{\text{s}} \) to represent coordinates where objects can be placed stably:
\begin{equation}
\mathcal{P}_{\text{s}} = \left\{ \mathbf{p}_i \in \mathcal{P} : \forall t, \lVert q(t, \mathbf{p}_i) - q(1, \mathbf{p}_i) \rVert^2 \leq \sigma^2 \right\}
\end{equation}
where \( q(1, \mathbf{p}_i) \) is the quaternion scalar component at \( t=1 \) for each point \( \mathbf{p}_i \), and \( \sigma^2 \) is the variance of \( \{q(t, \mathbf{p}_i)\}_{t=1}^{T}\).
More specifically, to determine the robustness of the placement stability, small perturbations are injected after the object has been placed. These perturbations can imitate real world conditions where external factors, such as vibrations or impacts, may affect the stability of the placed object.
If the objects can tolerate these perturbations, the robustness and real world applicability of placement tasks will be increased.

Specifically, for all $\mathbf{p} \in \mathcal{P}_{s}$, to compute the stability reward \( \mathrm{r}_{\text{s}}(\mathbf{p}) \) for each coordinate, we use a simple linear reward function as follows:
\begin{equation}
    \mathrm{r}_{\text{s}}(\mathbf{p}) = 100 \cdot (q_{\text{max}}(\mathbf{p}) - q_{\text{min}}(\mathbf{p}))
\end{equation}
where \( q_{\text{max}}(\mathbf{p}) \) and \( q_{\text{min}}(\mathbf{p}) \) are the maximum and minimum quaternion scalar components associated with point \( \mathbf{p} \), respectively.

\subsection{Context Understanding via Reasoning}
\label{common-sense-suitability}
Though the set of $\mathcal{P}_{\text{s}}$ points, that are verified in Sec~\ref{physics-stability}, are determined to be feasible, it may contain some options that do not take into account the context of the scene (e.g., putting a book on a desk instead of a bookshelf, putting an item in a different category when it should be categorized). 
Therefore, we aim to analyze the reasonableness within the limited range of $\mathcal{P}_{\text{s}}$ that corresponds to the current scene's situation and context.
An LLM~\cite{22_oh} uses the input of object labels from the same RGBD image in Sec. \ref{physics-stability} and generates one or more words representing the specific receptacle where the object could reasonably be located (i.e., reasonableness).

Since we have transferred the robot's ego-centric view from the real world to the simulator in Sec~\ref{physics-stability}, we can automatically identify the region where the coordinates of $\mathcal{P}_{\text{s}}$ are located.
% based on where the object belongs.
Based on the generated receptacle, \( \mathcal{P}_{\text{r}} \) is a set of points that are within 0.1 meters of the receptacle object, which is included in \( \mathcal{P}_{\text{s}} \).
For all \( \mathbf{p} \in \mathcal{P}_{\text{r}} \), the reward \( r_{\text{r}}(\mathbf{p})=+1 \) when the receptacle is a reasonable region, \( r_{\text{r}}(\mathbf{p})=0 \) otherwise.

% Define about Similarity.
\begin{figure}[t!] 
	\centering
	% First row
	\begin{minipage}{\linewidth}
		\centering
		\subfigure[]{
			\includegraphics[width=0.25\linewidth]{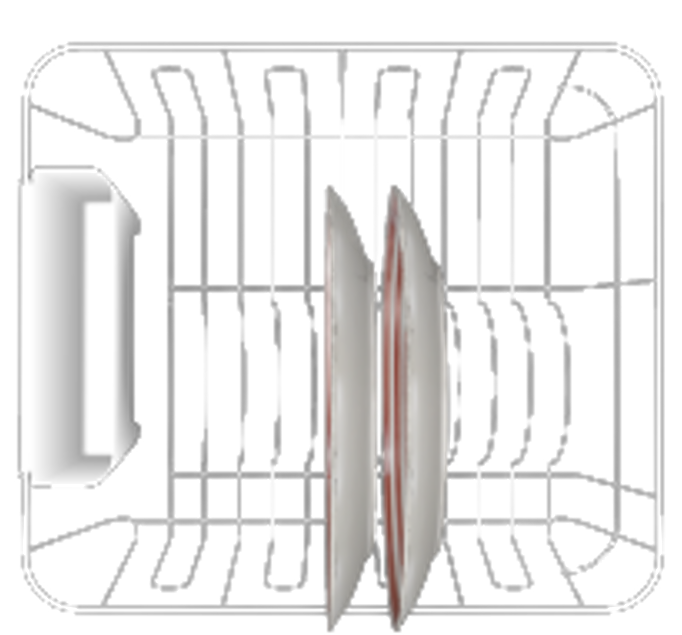}
			\label{white_dish_rack}
		}
		\subfigure[]{
			\includegraphics[width=0.25\linewidth]{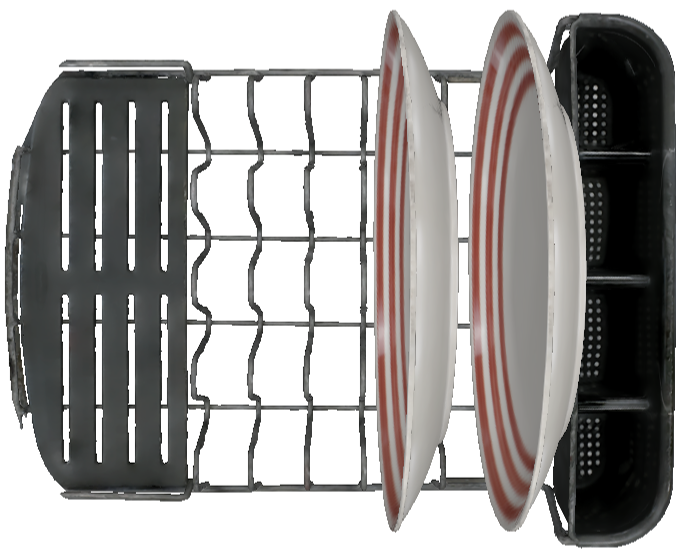}
			\label{black_dish_rack}
		}
		\subfigure[]{
			\includegraphics[width=0.25\linewidth]{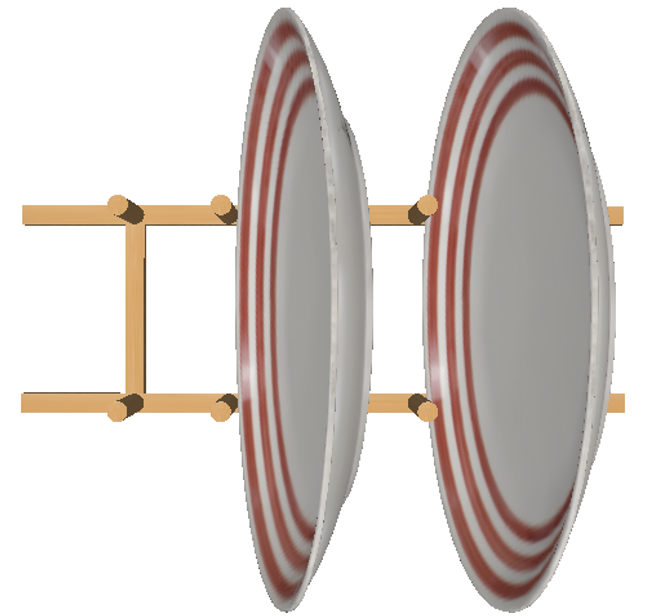}
			\label{small_dish_rack}
		}
	\end{minipage}
	
	% Second row
	\begin{minipage}{\linewidth}
		\centering
		\subfigure[]{
			\includegraphics[width=0.25\linewidth]{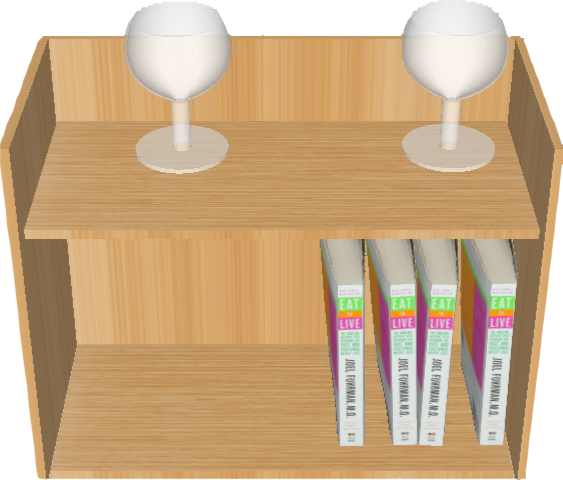}
			\label{bookshelf_normal}
		}
		\subfigure[]{
			\includegraphics[width=0.25\linewidth]{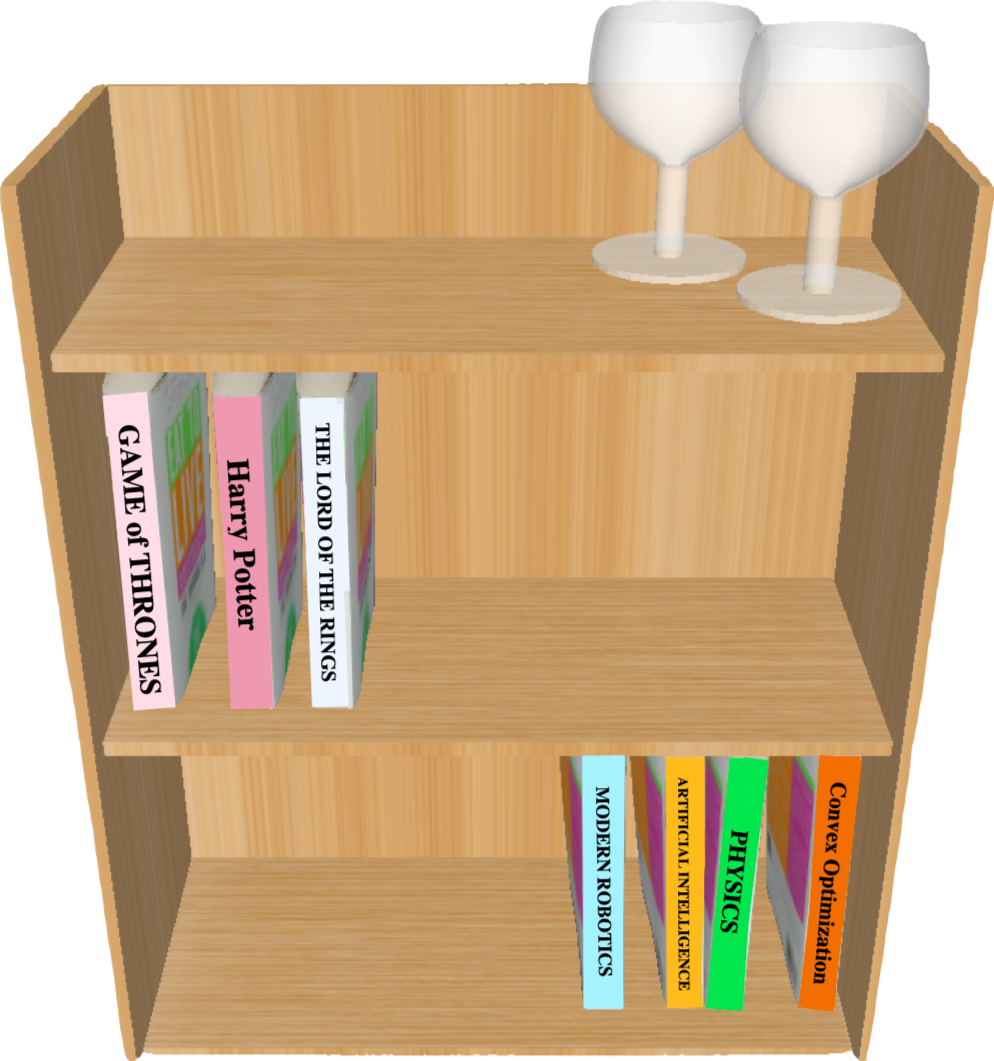}
			\label{bookshelf_dense}
		}
		\subfigure[]{
			\includegraphics[width=0.25\linewidth]{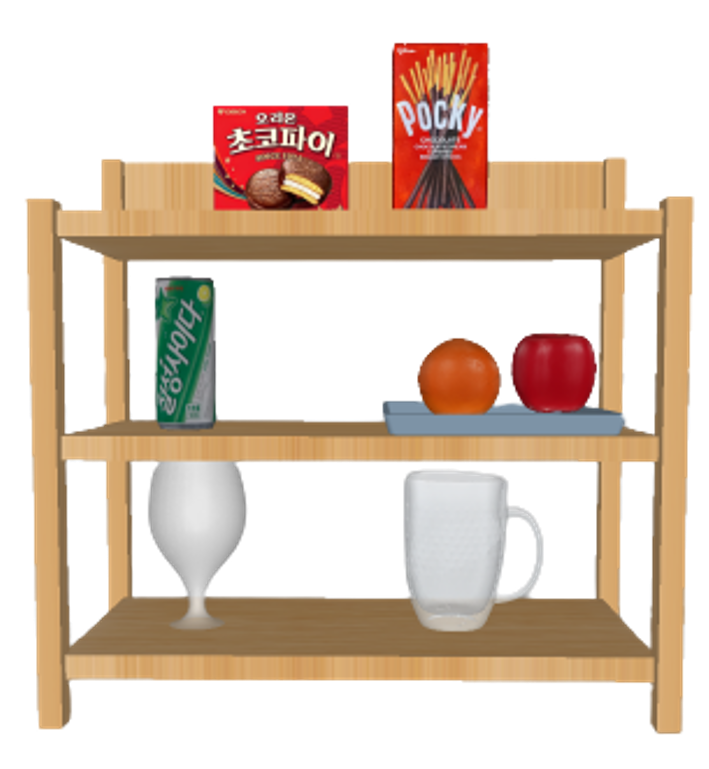}
			\label{sim_shelf}
		}
	\end{minipage}

        \caption{Randomization for the \textbf{Dish Rack} environment is (a) Small Gap, (b) Medium Gap, (c) Large Gap. Randomization for the \textbf{Bookshelf} environment is (d) Two-Tiered Bookshelf, (e) Three-Tiered Bookshelf. \textbf{Category} environment uses (f) Three-Tiered Shelf.}
	\label{fig:simulation_randomization}
\end{figure}

\subsection{\textbf{SPOTS}:~Reward-Guided Sampling}
With a set of rewards based on both the stability verification step in Sec.~\ref{physics-stability} and receptacle reasoning Sec.~\ref{common-sense-suitability}, the final procedure is to generate candidate place locations.
To characterize the distribution of feasible and suitable 3D points for object placement, we utilize a kernel density estimation (KDE) \cite{terrell_92_variable-kde}. Specifically, we use the radial basis function (RBF) kernel. The KDE for each point cloud is defined as:
\begin{equation}
    \hat{f}_h(\mathbf{{p}}, r(\mathbf{p})) = \frac{1}{N\cdot h} \sum_{\mathbf{p}_{i}\in \mathcal{P}} r(\mathbf{p}_{i}) \cdot K\left(\frac{\mathbf{p} - \mathbf{p}_{i}}{h}\right)
\end{equation}
where $N$ is the total number of coordinates, $h$ is the bandwidth parameter, $K$ is the kernel function, $\mathbf{p}$ is a random variable, $r(\mathbf{p}_{i})$ is the weight of the $i^{th}$ coordinate in $\mathcal{P}$, and $\{ \mathbf{p}_{i} \} ^N_{i=1}$, $\{r(\mathbf{p}_{i})\}^N_{i=1} $ are obtained in previous steps.

To emphasize the reward values for the receptacles within the set of physically stable coordinates \( \mathcal{P}_{s} \), we formulate the combined density distribution \( \mathcal{D}_{c} \) as follows:
\begin{equation}
    \mathcal{D}_{c}(\mathbf{p}) = \hat{f}_h(\mathbf{p}_{s}, r_{s}(\mathbf{p}) \cdot \beta + r_{r}(\mathbf{p}) \cdot (1 - \beta))
    % \mathcal{D}_{c}(\mathbf{p}_{s}) = \hat{f}_h(\mathbf{p}_{s}, r_{s}(\mathbf{p}) \cdot \beta + r_{r}(\mathbf{p}) \cdot (1 - \beta))
\end{equation}
where \( r_{s} \) and \( r_{r} \) represent the rewards associated with each point $\mathbf{p}_{s}$ and $\mathbf{p}_{r}$ in \( \mathcal{P}_{s} \) and \( \mathcal{P}_{r} \), respectively, and $\mathbf{p}$ is the union of $\mathcal{P}_{s}$ and $\mathcal{P}_{r}$. 
Importantly, sets \( \mathcal{P}_{r} \) and \( \mathcal{P}_{s} \) are mutually exclusive, enabling focused design for either physical stability or reasonableness without conflicting effects.
The parameter \( \beta \) is tunable and balances between physical stability and reasonableness. Although \( \beta \) is tunable, it was held constant at 0.1 in all experiments.

To summarize, based on contextual information from the input task description, such as user preferences and the current scene, our method generates a combined density $\mathcal{D}_{\text{c}}$ for candidate placements. This density not only accounts for the physical robustness and reasonableness of the placement task but also aligns with the likelihoods associated with the task description and receptacle. A higher $\mathcal{D}_{\text{c}}$ indicates a better fit with these factors. From this combined distribution, we sample probable placements, allowing the user to choose the one that maximizes our density function.

%
% Experiments
%
\section{Experiments}
% Objective of each experiment
We conduct experiments in both simulation and real world settings to accomplish distinct yet complementary goals. In simulation, our objective is to evaluate the effectiveness of our stability verification process.
Our real world experiments aim to assess $\textbf{SPOTS}$'s ability to accurately infer the appropriate location using both spatial and semantic reasoning via LLM. For comparison with other LLM-based baselines, we also measured the total number of input tokens and the time spent on interactions.

We designed a task that categorizes objects based on similarity. The reasoning criteria, termed \textbf{similarity}, varies for each experiment and serves as the ground truth for evaluating reasoning abilities.
To be more specific, we validate our algorithm on three scenarios. The scenario involves a) placing plates on a dish rack, b) placing books on a bookshelf (in simulation), and c) categorizing objects based on similarity (both in the simulation and real world).
The objects vary in size, shape, and position with each scenario.
Our evaluations are twofold: (1) we present $\textbf{SPOTS}$'s place stability in simulation environments (Section \ref{exp_place_stability}), and (2) we showcase the potential of $\textbf{SPOTS}$ in the semi-autonomous teleoperation framework (Section \ref{exp_realworld}) when reasoning capability is required (i.e., semantic reasoning or common knowledge).

%
% Environment Setup
%
\subsection{Environment setup}
Our real-to-sim transfer module, illustrated in Fig.\ref{fig:overview}, utilizes OWL-ViT \cite{owl_vit} for open-vocabulary object detection and AprilTags \cite{olson11tags_apriltag} for pose estimation, based on input from an RGBD vision sensor.
The detected objects form a label super-set that includes nine categories\footnote{\texttt{'DishRack', 'Bowl', 'BookShelf', 'Fruit', 'Beverage', 'Snack', 'Tray', 'Glass', 'Book'}}, for a total of 21 object assets. For each detected object, we assume the corresponding 3D asset is available. These assets are transferred into a simulation environment that mimics the real world as closely as possible. This reconstructed environment is the basis for all subsequent evaluations.
The framework is built on the MuJoCo \cite{mujoco} simulator, using assets from the YCB \cite{ycb_dataset} and Google Scanned dataset \cite{downs22google_google_scanned}. We use a tabletop manipulation framework with a 6-DoF robot arm and \texttt{gpt-3.5-turbo}~\cite{00_GPT}.

%
% Baselines
%
\subsection{Baselines \& Metrics}
We compare \textbf{SPOTS} to three prior methods: LLM-GROP~\cite{LLM-GROP}, Code-as-Policies (CaP)~\cite{CaP}, and Language-to-Reward (L2R)~\cite{Language-to-Reward}. LLM-GROP uses two different template-based prompts; one extracts semantic relationships with examples, and the other one predicts geometric spatial relationships for varying scene geometry. CaP generates policy code for the robot motion using a pre-defined low-level primitive function. 
L2R defines reward parameters that can be optimized, and the reward function is designed for moving a manipulator to a parameterized placement position.

Our evaluation metrics are the place stability and reasonableness of the suggested object placements. 
The stability success rate is based purely on the physical stability of object placement in simulations, whether that object is placed stable (i.e., Sta. S/R). 
Reasonableness success rate (i.e., Rea. S/R), on the other hand, is based on whether object placement aligns with the ground truth that we define. Evaluating reasonableness success criteria is manually designed; more details are described in Sec \ref{exp_place_stability}.
These metrics assess the overall effectiveness of placements in ensuring both stability and reasonableness.
These specific criteria are the ground truth for confirming appropriate locations in our experimental validation.
Furthermore, we measure the time taken for the inference and the number of input and output tokens to measure the efficiency of utilizing LLMs. 

%
% Simulation Results
%
\subsection{Simulation Experiments}
\label{exp_place_stability}

% Verified Coordinates
\begin{figure}[t]
    \centering
    \includegraphics[width=0.48\textwidth]{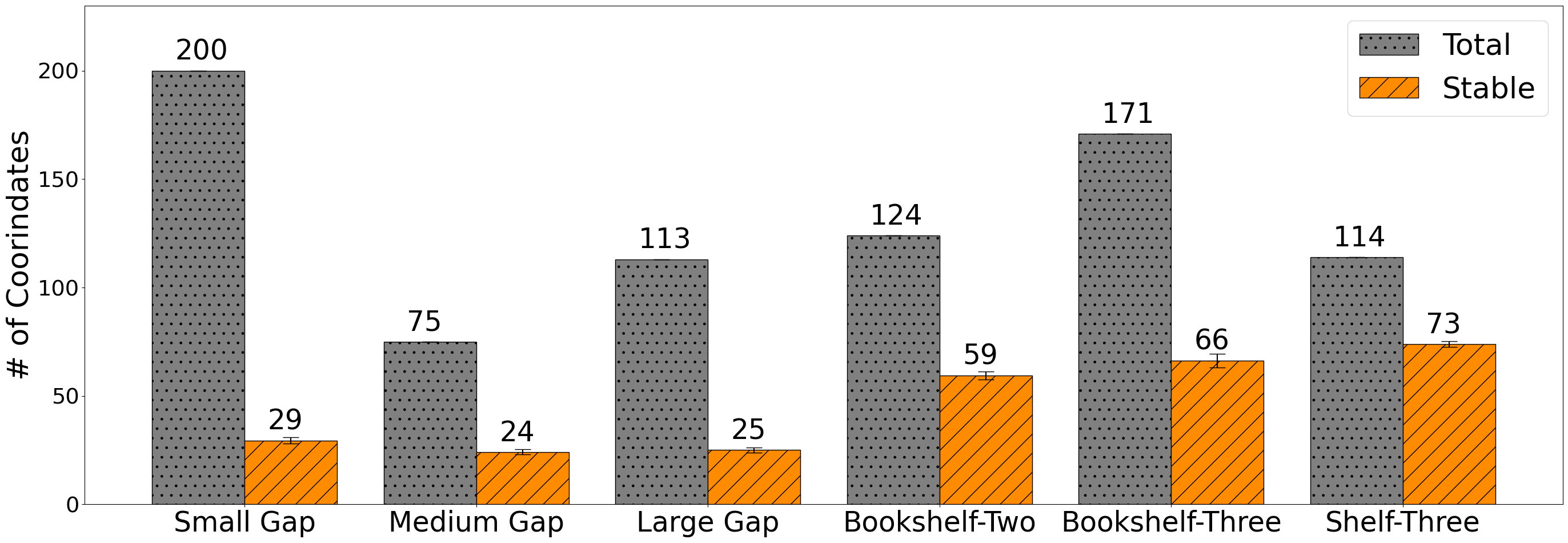}
    \caption{Result of stability verification module: Performed for all environments in our experiments, not including reasoning module. The ratio of stable coordinates to the total number of coordinates is very low. This indicates that the task we are assuming is physically difficult to be stably located.}
\label{fig:verification_result}
% \vspace{-0.1pt}
\end{figure}

We conducted experiments in three different scenarios within the simulation environment: 1) \textbf{Dish rack}, 2) \textbf{Bookshelf}, and 3) \textbf{Category} scenario. For each setting, we randomized the environment to evaluate the performance. \textbf{SPOTS} samples 10 coordinates per experiment and repeats a total of 20 times. The baseline model has 20 interactions. 
\paragraph{\textbf{Dish Rack}} On \textbf{Dish Rack} experiments, we focused solely on evaluating the place stability of objects, assuming that the receptacle locations are known and accurate. Baseline models were also evaluated under the same conditions for a fair comparison.
There were three different types and sizes of racks used for randomization. 
From Fig.~\ref{fig:verification_result}, we can infer that the number of feasible coordinates in the \textbf{Dish Rack} scenario is small and that this task is difficult to perform without physical simulation.
From Table \ref{exp_simulation}, the success rate for all the compared methods is over $60\%$, when highly detailed spatial information is given. 
The \textbf{SPOTS} outperforms the compared method with a margin of $0.05$ on the success rate, indicating that the proposed method can robustly place the object in a complex environment. 

\paragraph{\textbf{Bookshelf}} The \textbf{Bookshelf} experiment assesses reasoning ability and consists of two scenes. 
In the two-tiered bookshelf environment, books are on the first (lower) tier, wine glasses are on the second (upper) tier, and we define the reasonable location as the first tier of the bookshelf. In a three-tiered bookshelf, the first and second tiers each contain books of the same genre (e.g., textbook, novel, fairy-tale), and the third tier contains a wine glass. In this case, the tier that matches the genre with the placement object is denoted as ground truth. 
The experimental results on Table \ref{exp_simulation} show that \textbf{SPOTS}
outperforms the baselines in Sta. S/R and Rea. S/R with a gap of $0.05$ and $0.00$, respectively. While CaP~\cite{CaP} is observed to have comparable performance in two-tiered bookshelf scenarios, \textbf{SPOTS} reduces the number of prompts needed from $985$ to $323$. This disparity arises because CaP requires carefully tuned few-shot prompts, whereas \textbf{SPOTS} can execute the task without needing precise explanations or hand-crafted examples to perform the task.

\paragraph{\textbf{Category}}Finally, we designed a task where objects were categorized based on \textbf{similarity}. We define three types of similarity here: color, object property, and shape, which serve as the ground truth for evaluating reasoning abilities.
The reasonable receptacle changes based on this \textbf{similarity} type (i.e., the ground truth of the Rea. S/R changes based on the similarity). 
The scenario consists of a three-tiered shelf, each with different objects (e.g., glass cup, beverage, snack, fruit). 
In Table \ref{exp_simulation}-\textbf{Category}, \textbf{SPOTS} surpasses all compared methods by a margin of $0.15$ and $0.10$ in terms of Sta. S/R and Rea. S/R, respectively. 
By separating the tasks of predicting receptacles and ensuring physical robustness into two distinct modules, we find that \textbf{SPOTS} achieves a higher success rate while using fewer tokens compared to the methods that enforce LLMs to predict both robotic plans while understanding the context.
From this experiment, we would like to posit that \textbf{SPOTS} has great capability of promptable placement tasks, which considers both physically stable and reasonable regions, and \textbf{SPOTS} has a good distribution, where reasonable positions can be sampled.

\begin{table*}[]
\resizebox{\textwidth}{!}{%
    \begin{tabular}{clcccccccccccc}
    \hline
    \multicolumn{14}{c}{Simulation}                                                                                                                                                                                                                        \\ \hline
    \multicolumn{2}{c|}{\multirow{3}{*}{Model}} & \multicolumn{12}{c}{\textbf{Dish Rack}}                                                                                                                                                                        \\ \cline{3-14} 
    \multicolumn{2}{c|}{}                       & \multicolumn{4}{c|}{Small Gap Dish Rack, Fig.~\ref{fig:simulation_randomization}a}                                                            & \multicolumn{4}{c|}{Medium Gap Dish Rack, Fig.~\ref{fig:simulation_randomization}b}                                              & \multicolumn{4}{c}{Large Gap Dish Rack, Fig.~\ref{fig:simulation_randomization}c}                         \\ \cline{3-14} 
    \multicolumn{2}{c|}{}                       & \multicolumn{2}{c}{Time}               & S/R      & \multicolumn{1}{c|}{\# of token} & \multicolumn{2}{c}{Time} & S/R      & \multicolumn{1}{c|}{\# of token} & \multicolumn{2}{c}{Time} & S/R      & \# of token \\ \hline
    \multicolumn{2}{c|}{LLM-GROP~\cite{LLM-GROP}}               & \multicolumn{2}{c}{$1.78\pm0.29$}               & 0.80    & \multicolumn{1}{c|}{231}        & \multicolumn{2}{c}{$1.92\pm0.43$}     & 0.80         & \multicolumn{1}{c|}{253}           &\multicolumn{2}{c}{$1.72\pm0.35$}     & 0.9       &  244          \\
    \multicolumn{2}{c|}{CaP~\cite{CaP}}                    & \multicolumn{2}{c}{$25.32 \pm 5.91$}              & 0.60     & \multicolumn{1}{c|}{823}        & \multicolumn{2}{c}{$13.97 \pm 1.12$}     & 0.65         & \multicolumn{1}{c|}{792}           & \multicolumn{2}{c}{$12.46\pm4.05$}     & 0.75         & 949           \\
    \multicolumn{2}{c|}{L2R~\cite{Language-to-Reward}}                    & \multicolumn{2}{c}{$5.16\pm0.35$}             &0.70      & \multicolumn{1}{c|}{1453}        & \multicolumn{2}{c}{$4.58\pm0.44$}     &0.70          & \multicolumn{1}{c|}{1306}           & \multicolumn{2}{c}{$5.21\pm0.24$}     &0.80          & 1454           \\
    \multicolumn{2}{c|}{\textbf{SPOTS (Ours)}}  & \multicolumn{2}{c}{$3.70 \pm 0.04$} & \textbf{0.85}     & \multicolumn{1}{c|}{-}          & \multicolumn{2}{c}{$1.48 \pm 0.04$}     & \textbf{0.85}         & \multicolumn{1}{c|}{-}           & \multicolumn{2}{c}{$5.41 \pm 0.02$}     & \textbf{0.95}         &  {-}          \\ \hline
    
    \multicolumn{2}{c|}{\multirow{3}{*}{Model}} & \multicolumn{8}{c|}{\textbf{Bookshelf}}                                                                                                    & \multicolumn{4}{c}{\textbf{Category}}         \\ \cline{3-14} 
    \multicolumn{2}{c|}{}                       & \multicolumn{4}{c|}{Two-Tiered Bookshelf, Fig.~\ref{fig:simulation_randomization}d}                                                     & \multicolumn{4}{c|}{Three-Tiered Bookshelf, Fig.~\ref{fig:simulation_randomization}e}                                    & \multicolumn{4}{c}{Three-Tiered Shelf, Fig.~\ref{fig:simulation_randomization}f}                    \\ \cline{3-14} 
    \multicolumn{2}{c|}{}                       & Time               & Sta. S/R    & Suit. S/R & \multicolumn{1}{c|}{\# of token} & Time & Sta. S/R & Suit. S/R & \multicolumn{1}{c|}{\# of token} & Time  & Sta. S/R    & Suit. S/R & \# of token \\ \hline
    \multicolumn{2}{c|}{LLM-GROP~\cite{LLM-GROP}}               & $1.90\pm0.45$               & 0.65           & 0.80        & \multicolumn{1}{c|}{428}        & $2.90\pm0.17$  & 0.65    & 0.80     & \multicolumn{1}{c|}{658}        &$1.95\pm0.61$       & 0.70           & 0.75         & 451            \\
    \multicolumn{2}{c|}{CaP~\cite{CaP}}                    & $6.98 \pm 0.22$               & \textbf{0.95}           & \textbf{0.95}         & \multicolumn{1}{c|}{985}        & \multicolumn{1}{c}{$6.87 \pm 1.80$}     & 0.85        & 0.65         & \multicolumn{1}{c|}{1168}           & $4.42\pm0.26$      & 0.70           & 0.6         & 1423           \\
    \multicolumn{2}{c|}{L2R~\cite{Language-to-Reward}}                    & $6.24\pm0.87$            &  0.75          & 0.80         & \multicolumn{1}{c|}{1153}        & $7.78\pm1.04$     & 0.8     & 0.80     & \multicolumn{1}{c|}{1493}      & $7.63\pm1.63$       &  0.75          & 0.75       & 1205           \\
    \multicolumn{2}{c|}{\textbf{SPOTS (Ours)}}  & $3.77 \pm 0.03$             & 0.90 & 0.95     & \multicolumn{1}{c|}{323}        & $4.50 \pm 0.01$ & \textbf{0.90}    & \textbf{0.85}     & \multicolumn{1}{c|}{430}        & $9.07 \pm 0.10$ & {\textbf{0.85}} & \textbf{0.85}     & 342        \\ \hline
    \end{tabular}%
}
\caption{Simulational Result: Experiments with six scenarios for a total of 3 environments. The \textbf{Dish Rack} scene was set up with different shapes and sizes of the racks. The \textbf{Bookshelf} scene was varied with diverse shelf sizes, positions, and genres. The \textbf{Category} scene was randomized using \textbf{similarity} criteria.}
\label{exp_simulation}
\end{table*}

\subsection{Real World Demonstration}
\label{exp_realworld}

\begin{table}[]
    \centering
    \resizebox{0.49\textwidth}{!}{%
    \begin{tabular}{ccccccc}
        \hline
        \multicolumn{7}{c}{\textbf{Realworld}} \\ \hline
        \multicolumn{1}{c|}{\multirow{3}{*}{Model}} & \multicolumn{6}{c}{\textbf{Tray}} \\ \cline{2-7} 
        \multicolumn{1}{c|}{} & \multicolumn{3}{c|}{Shape} & \multicolumn{3}{c}{Object Property} \\ \cline{2-7} 
        \multicolumn{1}{c|}{} & Time & S/R & \multicolumn{1}{c|}{\# of Token} & Time & S/R & \# of Token \\ \hline
        \multicolumn{1}{c|}{SPOTS (Ours)} & $3.58 \pm 0.08$ & 0.80 & \multicolumn{1}{c|}{362} & $5.69 \pm 0.04$ & 0.80 & 405 \\ \hline
    \end{tabular}%
}
\caption{Realworld Result: Experiments with the shape and object property as \textbf{similarity}.}
\label{exp_realworld}
\vspace{-0.2cm}
\end{table}

\begin{figure*}[t!] 
	\centering
	\subfigure[]{
	\includegraphics[width=0.3\linewidth]{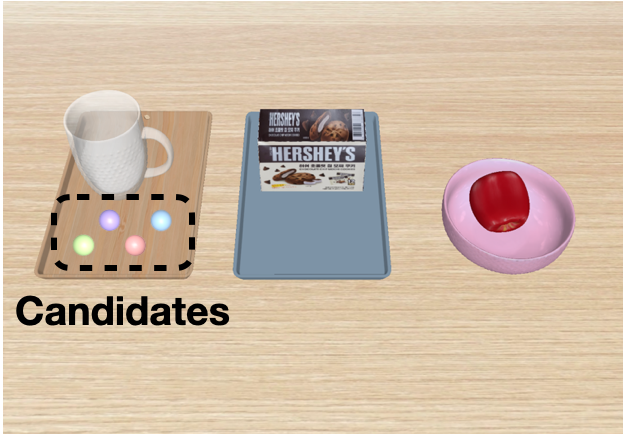}
	}
	\subfigure[]{
	\includegraphics[width=0.3\linewidth]{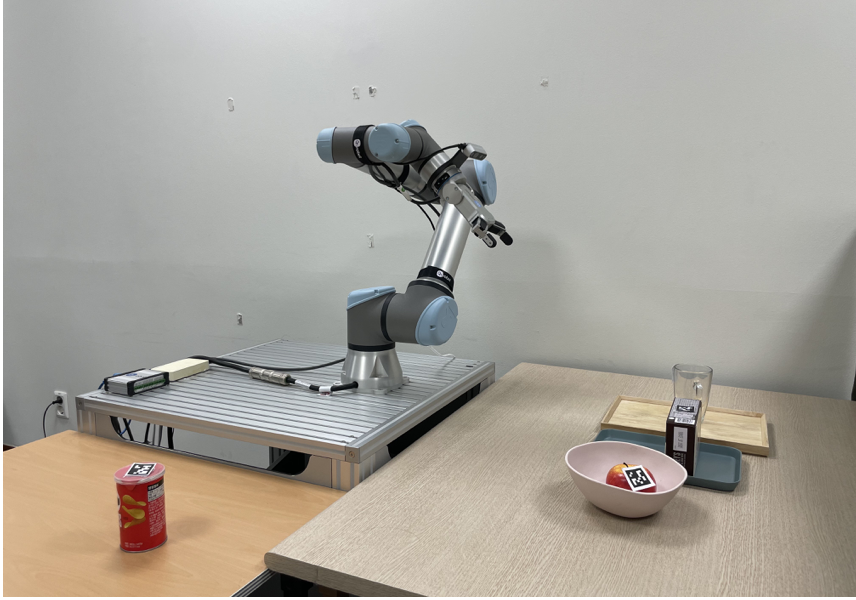}
	}
	\subfigure[]{
	\includegraphics[width=0.3\linewidth]{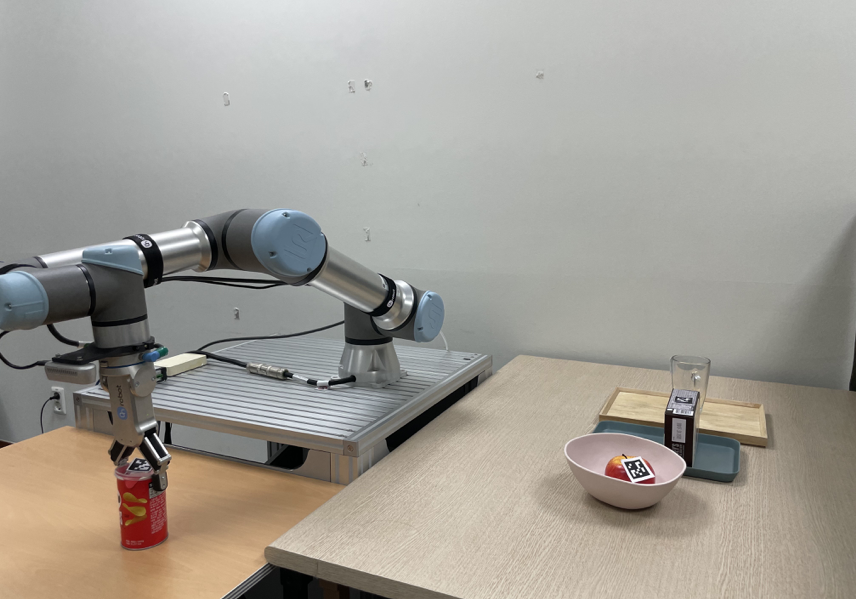}
	}
	\subfigure[]{
	\includegraphics[width=0.3\linewidth]{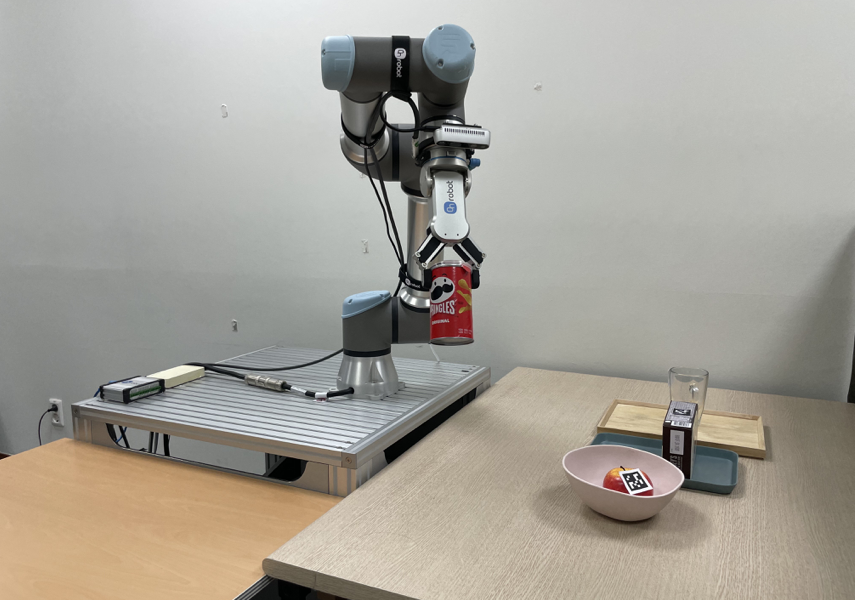}
	}
	\subfigure[]{
	\includegraphics[width=0.3\linewidth]{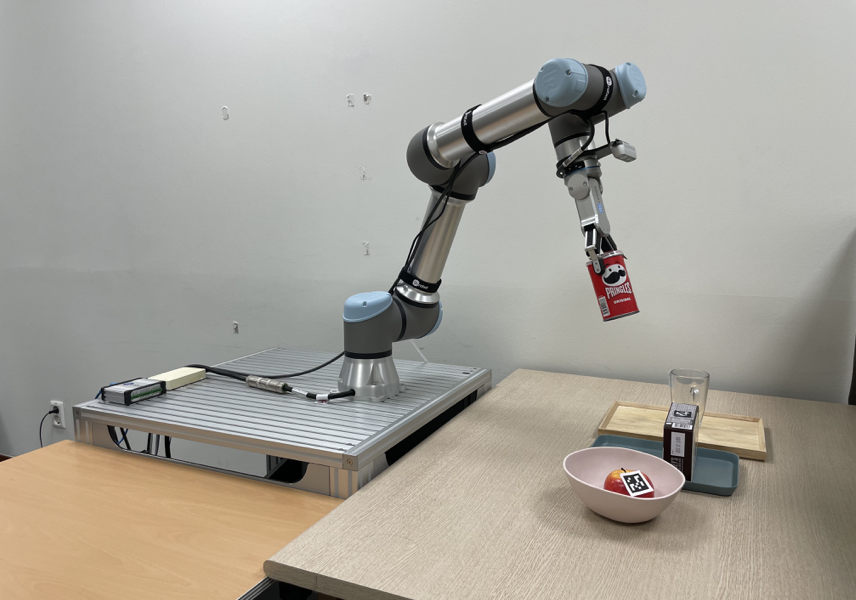}
	}
	\subfigure[]{
	\includegraphics[width=0.3\linewidth]{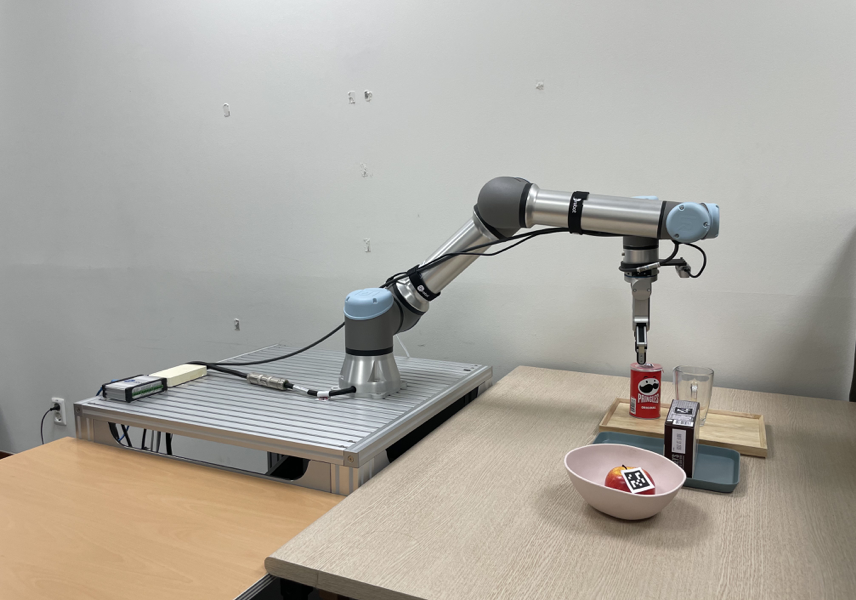}
	}
	\caption{Snapshots of real world robot experiments. Snapshots of real robot experiments: (a) shows the user interface from the perspective of the user, (b)-(f) show the robot performs a promptable placement task based on shape \textbf{similarity}.}
	\label{fig:snapshot}
\end{figure*}

In this experiment, we consider a scene with three different trays placed on a desk, each containing a single object. The target object for placement is a potato snack. Each type of similarity was evaluated five times, and performance was measured using the overall success rate (i.e., both Sta. S/R and Rea. S/R).
In both the shape and object property cases, the success rate was $0.8$, indicating a generally effective reasoning process. The failure cases occurred when incorrect matches were made due to incorrect reasoning.
From this experiment, we insist that the reasonable place varies depending on the task description given as input. Furthermore, we are able to accurately determine the stable positions to place the objects by reconstructing the robot's ego-centric view with the real-to-sim method. The $\beta$ term also allowed us to control the relationship between the physically stable and semantically reasonable locations.

% Limitation of Method
\subsection{Limitations}
Since our model utilizes LLMs, we need a language projection phase, which requires an accurate open vocabulary object detection model as well as knowing the superset of possible objects in the scene. 
This may hinder the usage of \textbf{SPOTS} in a novel environment. 
Moreover, for the sake of real-to-sim, the current version of our implementation requires CAD models to reconstruct the scene in a physics-based simulator. 
However, this restriction may be alleviated by having more accurate 3D sensing devices or using implicit representations such as NeRF2Real~\cite{byravan2022nerf2real_mujoco}. 

%
% Conclusion
%
\section{Conclusion}
The complexity of the ``place'' aspect of the pick-and-place robotic task emphasizes the need for systems that can integrate physical stability with semantic reasonableness. 
Our approach, $\textbf{SPOTS}$, uses physics-based simulation and semantic reasoning to achieve the optimal placement of objects in complex environments. 
The results show a reliable and situational system that bridges the difference between absolute robotic independence and human-like discrimination.
However, despite the great potential demonstrated by \textbf{SPOTS}, there is still a further need to improve its real-time flexibility and effectiveness, especially in more diverse situations.
Further development will concentrate on enhancing the incorporation of dynamic feedback from the environment and broadening the system's database. 
Additionally, incorporating user feedback loops can provide possibilities for continual improvement, aiding the growth of extremely responsive and adaptable robotic systems.

%%%%%%%%%%%%%%%%%%%%%%%%%%%%%%%%%%%%%%%%%%%%%%%%%%%%%%%%%%%%%%%%%%%%%%%%%%%%%%%%
\bibliographystyle{IEEEtran}
\bibliography{references}  % .bib

\end{document}